\documentclass[preprint]{vgtc}               

\ifpdf
  \pdfoutput=1\relax                   
  \usepackage{graphicx}                
  \DeclareGraphicsExtensions{.pdf,.png,.jpg,.jpeg} 
\else
  \usepackage{graphicx}                
  \DeclareGraphicsExtensions{.eps}     
\fi%

\graphicspath{{figures/}{pictures/}{images/}{./}} 

\usepackage{microtype}                 
\PassOptionsToPackage{warn}{textcomp}  
\usepackage{textcomp}                  
\usepackage{mathptmx}                  
\usepackage{times}                     
\usepackage{cite}                      
\usepackage{tabu}                      
\usepackage{booktabs}                  

\usepackage{caption}
\usepackage{subcaption}
\usepackage{enumitem}

\onlineid{1103}

\vgtccategory{Research}

\vgtcinsertpkg

\ieeedoi{10.1109/VIS54172.2023.00045}

\title{Visual Validation versus Visual Estimation: \\A Study on the Average Value in Scatterplots}

\newcommand*\samethanks[1][\value{footnote}]{\footnotemark[#1]}
\author{Daniel Braun\thanks{e-mail: \{braun\,$|$\,landesberger\}@cs.uni-koeln.de}\\ %
        \scriptsize University of Cologne %
\and Ashley Suh\thanks{e-mail: \{ashley.suh\,$|$\,remco.chang\}@tufts.edu}\\ %
     \scriptsize Tufts University %
\and Remco Chang\samethanks[2]\\ %
     \scriptsize Tufts University %
\and Michael Gleicher\thanks{e-mail: gleicher@cs.wisc.edu}\\ %
     \parbox{1in}{\scriptsize \centering University of \\ Wisconsin - Madison} %
\and Tatiana von Landesberger\samethanks[1]\\ %
     \scriptsize University of Cologne}

\teaser{
  \centering
     \begin{subfigure}[t]{0.32\textwidth}
         \centering
         \includegraphics[width=0.9\textwidth]{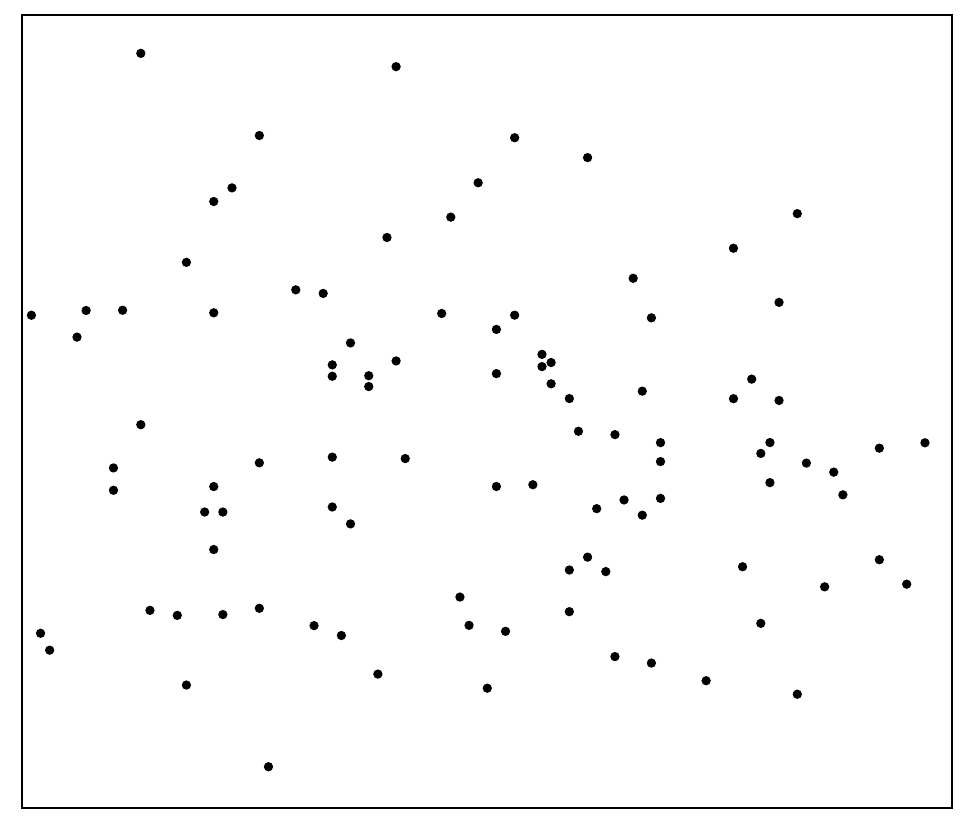}
         \caption{Scatterplot as shown in the visual estimation tasks.}
         \label{fig:scat_init}
     \end{subfigure}
     \hfill
     \begin{subfigure}[t]{0.32\textwidth}
         \centering
         \includegraphics[width=0.9\textwidth]{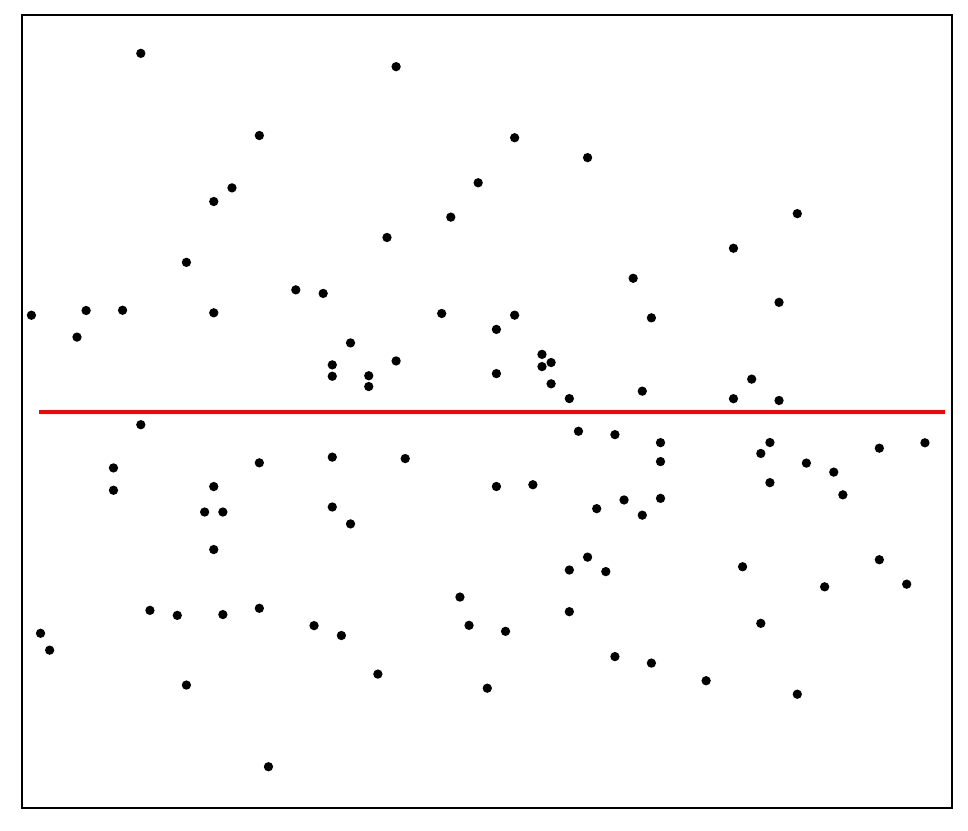}
         \caption{Scatterplot as shown in the visual validation tasks.}
         \label{fig:scat_line}
     \end{subfigure}
     \hfill
     \begin{subfigure}[t]{0.32\textwidth}
         \centering
         \includegraphics[width=0.9\textwidth]{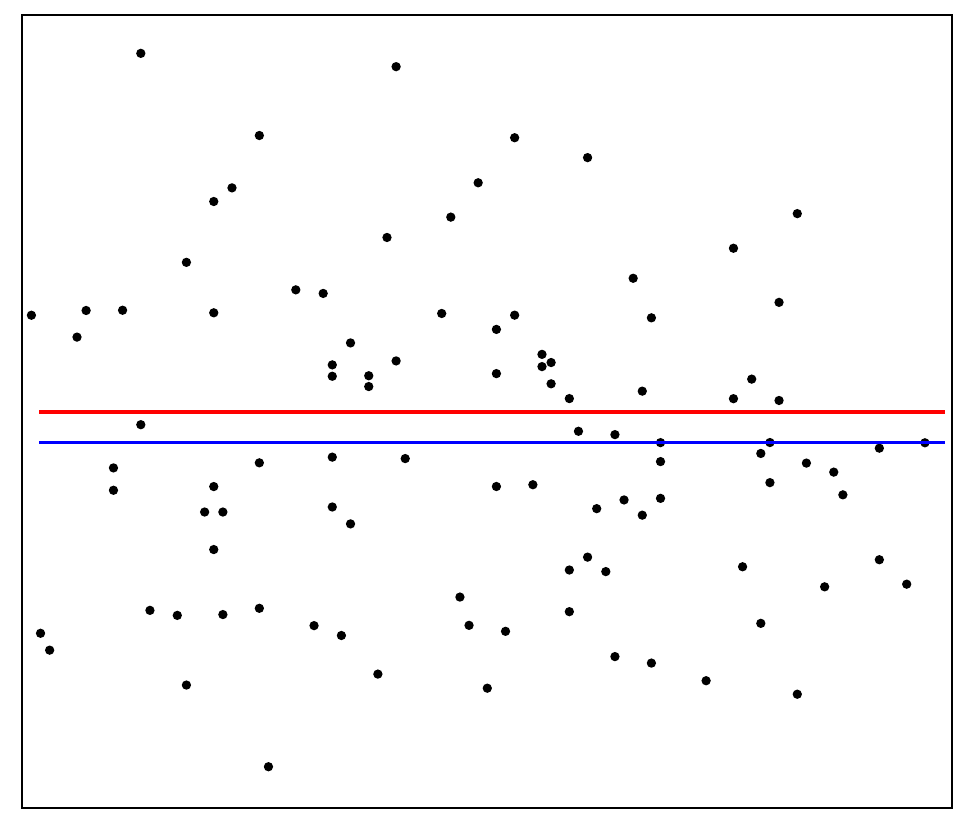}
         \caption{Scatterplot with the  true average value of the dots.}
         \label{fig:scat_c}
     \end{subfigure}%
        \caption{Example scatterplot shown to participants in our user study to investigate the differences between (a) visually estimating and (b) visually validating the average value of the shown data. In (c), the red lines indicate the upper border of the statistical 95\% confidence interval (CI) of the average value and the blue line shows the data's true average value.}
        \label{fig:teaser}
}

\abstract{%
  We investigate the ability of individuals to visually validate statistical models in terms of their fit to the data. While visual model estimation has been studied extensively, visual model validation remains under-investigated. It is unknown how well people are able to visually validate models, and how their performance compares to visual and computational estimation. As a starting point, we conducted a study across two populations (crowdsourced and volunteers). Participants had to both visually estimate (i.e, draw) and visually validate (i.e., accept or reject) the frequently studied model of averages. 
  Across both populations, the level of accuracy of the models that were considered valid was lower than the accuracy of the estimated models. 
  We find that participants' validation and estimation were unbiased. Moreover, their natural critical point between accepting and rejecting a given mean value is close to the boundary of its 95\% confidence interval, indicating that the visually perceived confidence interval corresponds to a common statistical standard.
  Our work contributes to the understanding of visual model validation and opens new research opportunities.}

\CCScatlist{
  \CCScatTwelve{Perception}{Visual model validation}{Visual model estimation}{User study}; \CCScatTwelve{Information visualization}{}{}{}
}

\usepackage{xcolor}
\definecolor{mred}{rgb}{.80,.12,.30}
\definecolor{grey}{rgb}{.5,.5,.5}
\definecolor{turquoise}{rgb}{.04,.79,.93}

\newif\ifnotes
\notestrue

\usepackage[normalem]{ulem}
\usepackage{soul}
\usepackage{hyperref}

\let\origcite\cite
\renewcommand{\cite}[1]{\ifnotes\mbox{\origcite{#1}}\else \origcite{#1}\fi}

\begin{document}

\setlength\abovedisplayskip{5pt}
\setlength\belowdisplayskip{5pt}


\firstsection{Introduction}

\maketitle


In today's data-driven world, individuals with a range of statistical expertise are tasked with visually validating statistical models fitted to data for a variety of purposes. 
The general public 
use visual model validation when consuming news media visualizations, e.g., to become informed about changes of COVID-19 cases over time (such as using a 14-day moving average model visualization\cite{dansana2020global}).
Similarly, domain experts  
also use visual validation to quickly determine whether a particular model is a good fit to the underlying data (e.g., meteorologists validating a model for assimilation).

To perform model validation, an individual checks that a given model result is a good fit to the data. This model verification can be done computationally or visually; however, statistics computed to validate models are often insufficient to fully describe the underlying data and model's quality\cite{Matejka.2017}.
An example is Simpson's paradox, where the evaluation of groups differ depending on whether or not they are divided into subgroups\cite{Simpson.1951, Blyth.1972}.
Thus, it is essential in practice to validate computed models not only through statistical tests, but also visually to ensure that they are accurate and reliable.

Despite the prevalence of \textit{visual model validation}\cite{Muehlbacher.2013}, there is markedly little research done to understand viewers' ability to perform these processes.
Instead, most research in the perception of statistical modeling has focused on \textit{model estimation}\cite{Correll.2017, Gleicher.2013, Hong.2022, Rensink.2010, Xiong.2022, Yuan.2019, Yang.2019, Harrison.2014, Kay.2016, Newburger.2022}, that is, the ability of an individual to perceive, draw, or predict a model that is appropriately fitted to the data. While these studies help inform our understanding of experts and non-experts' strengths and limitations in \textit{estimating} models, they do not advise us of an individual's ability to \textit{validate} how well a statistical model fits the data. 
Thus, it is currently unknown how well people validate models in comparison to their estimation ability.

We seek to establish a baseline for understanding the differences between an individual's ability to perform model estimation and validation.
As an initial exploration, we chose to use \textit{averages} (mean expected value) in scatterplots. The model of averages is a simple but generalizable model that is widely used in visual analysis applications\cite{Demir.2016, Dube.2019}. It corresponds to the result of a linear regression to a constant that summarises the underlying data and provides a simplified representation of its central tendency. It also allows us to compare our results against existing literature on visual model estimation\cite{Gleicher.2013, Hong.2022, Yuan.2019} and to follow recognized practice of ``finding trends'' in scatterplots\cite{Munzner.2014}.

Using this setting, we compare the visual processes of model validation and estimation and answer the following research questions:
\begin{itemize}[noitemsep, nolistsep]
    \item \textbf{RQ1}: Are individuals able to perform visual validation consistently and without bias for averages in scatterplots? 
    \item \textbf{RQ2}: How does performance in visual validation relate to the accuracy of visual estimation in scatterplots?
\end{itemize}

We conducted a human subjects study with two participant groups to address these research questions. Using a between-subjects study design, participants either had to validate whether a shown line in a scatterplot represents the average value of the dots (\autoref{fig:scat_line}) or had to estimate the average line on their own (\autoref{fig:scat_init}).
Our study finds that the participants are consistent and unbiased (i.e. not influenced by data distribution) when deciding the threshold for a model to be accepted as valid. Furthermore, the required level of accuracy for validation was found to be lower than what can be estimated through visual inspection.
We find that the critical point between accepting and rejecting a line is close to the boundary of the 95\% confidence interval (CI) of the true mean value (i.e., a common statistical criteria for validity). 
These results have two major implications.
First, viewers accept models that are less accurate than they can estimate visually.
Second, individuals' ability to judge a model within the 95\% CI suggests that the visual perceived confidence interval corresponds to a common statistical standard.

Overall, our work contributes towards a new understanding of how well people can visually validate a model when given data. 
Our study 
provides a baseline for contributing towards this understanding, opening up future work for generalization beyond scatterplots and the model of averages.

\section{Related Work}
\label{sc:rw}

\paragraph{Visual Model Estimation}

Research on visual model estimation has investigated the concept of ensemble coding, the rapid extraction of visual statistics on distributions of spatial or featural visual information to estimate actual statistics on data\cite{Szafir2016ensemble}. In particular, the estimation of linear regression to a constant (i.e., average) has been extensively studied for scatterplots: Hong et al. explored the influence of a third dimension, encoded by color or size of the dots, on mean estimation\cite{Hong.2022}. They found that people's mean position estimations are biased towards larger and darker dots. Similarly, Gleicher et al. investigated the perception of average dot height in multi-class scatterplots and found that the perceptual process is robust against more points, sets, or conflicting encodings\cite{Gleicher.2013}. Additionally, Yuan et al. found that the visual estimation of averages depends on the visual encoding of the data and that people use primitive perceptual cues for their estimation\cite{Yuan.2019}. In line with these previous works, this paper uses scatterplots and averages as a baseline for visual model validation research.

Visual estimation of linear trends, e.g., correlations of two-dimensional data, has also been considered from several aspects. Rensink and Baldridge examined the statistical properties at which a correlation can be perceived for data in scatterplots\cite{Rensink.2010}. Rather than statistical properties, Yang et al. focused on visual data patterns in scatterplots and their effect on the perception of correlations\cite{Yang.2019}. Their results suggest that visual features, such as bounding boxes, influence people's correlation judgments. Xiong et al. consider the correlation in scatterplots from a data semantics point of view and found that people estimate the correlation more accurately with generic axis labels than with meaningful labels\cite{Xiong.2022}. 
Comparisons and rankings of different correlation visualizations based on Weber's law showed measurable differences between various designs (with scatterplots being the best) and that the performances differ significantly for positive and negative correlations\cite{Harrison.2014, Kay.2016}. These works support our choice of scatterplots for our study
and their findings influenced our hypotheses and stimuli design. 

Research has also been conducted on the visual estimation of more complex models. 
For example, Correll and Heer investigated people's ability to perform ``regression by eye'' for different types of trends and visualization types\cite{Correll.2017}. Their results showed that individuals can accurately estimate trends in many standard visualizations; however, both visual features and data features (e.g., outliers) can affect the results. Newburger et al. focused specifically on fitting bell curves to different visualization types\cite{Newburger.2022}. They found that people are accurate at finding the mean, but tend to overestimate the standard deviation. We aim to explore whether the findings from these papers also apply to visual validation for less complex models.

\paragraph{Visual Model Validation}

The literature on visual model validation is currently limited. 
Correll et al. evaluated scatterplots, histograms and density plots as means to support data quality checks\cite{Correll.2018}. Their findings suggest that problems arise as soon as overplotting occurs, which informed the design of the scatterplots in our study.

Visual model validation is particularly significant in machine learning. Chatzimparmpas et al. provide an overview of how visualization is currently used to interpret machine learning models\cite{Chatzimparmpas.2020}. Mühlbacher and Piringer present a partition-based framework for creating and validating regression models that combines the use visualizations with a relevance measure for ranking features\cite{Muehlbacher.2013}. Our work aims to develop an understanding of visual model validation which can inform the development of future visualization and machine learning systems.

Model validation is closely related to the viewers' trust in these models and their visualizations. In their state-of-the-art report, Chatzimparmpas et al. summarize the importance of using visualizations to increase trust in machine learning models\cite{Chatzimparmpas.2020a}. In addition to machine learning, the relationship between visual design and trust is an important area of research that should be considered in future work on visual validation\cite{Rogowitz.1996, Peskov.2020, Mayr.2019, Elhamdadi.2022, Dasgupta.2017}.

\section{User Study}
\label{sc:userstudy}

To address our research questions, we first conducted an exploratory pilot study. Based on the results of the pilot, we formulated hypotheses that were subsequently tested in a confirmatory main study. 

\subsection{Experimental Design}
\label{ssc:expdesign}

The same experimental design and structure was used for both the pilot study and the main study.
We used two \textbf{tasks} to compare the following visual processes:
\begin{itemize}[noitemsep, nolistsep]
    \item \textit{Visual estimation}: Participants had to draw a horizontal line in the plot by hovering over the image with the mouse and clicking on the desired position on the y-axis (\autoref{fig:scat_init}).
    \item \textit{Visual validation}: Participants were shown a scatterplot with a line already drawn and were asked to indicate whether the line was ``too high,'' ``too low,'' or ``about the same'' in relation to their perceived true mean value of the dots (\autoref{fig:scat_line}).
\end{itemize}
We chose a \textbf{between-subject design} of our study to prevent learning effects between the tasks and to reduce the number of trials per participant\cite{Charness.2012}. To maintain consistency, we used the same \textbf{plots} for both the validation and estimation trials.
Each scatterplot, which had dimensions of $100 \times 100$ pixels, contained 100 data points uniformly distributed on the x-axis. The only difference between the two between-subject groups was the lines shown in the validation tasks.

For the analysis, we defined the deviation of a line based on the confidence interval \textit{(CI)} calculation, quantifying the \textit{deviation} from the true mean as a proportion of the standard deviation:
\begin{equation}
    \mbox{shown value} = \mbox{true mean} + \mbox{\textit{deviation}} \cdot \mbox{standard deviation}
\end{equation}

Using regression calculation, the expected acceptance range for lines falls within the 95\% CI.
Given our fixed number of data points and distributions, the 95\% CI is set at a deviation of $0.198$. Thus, all lines with a lower deviation should be considered acceptable.

The \textbf{study procedure} began with a training phase, followed by the experimental study. Each page of the study interface displayed one trial (i.e., one plot) and response times were recorded. A display size of at least 13'' was recommended. The order of the trials (i.e., the order in which participants saw different deviations) was randomized to minimize learning effects. At the conclusion of the study, participants were asked to describe their strategies for completing the tasks and rate the task's difficulty on a 5-point Likert scale\cite{Vagias.2006}.

\subsection{Pilot Study}
\label{ssc:pilotstudy}

To gain initial understanding of visual validation and its differences from visual estimation, we conducted a pilot study with 12 participants (7 for validation and 5 for estimation). Each participant answered 30 trials with pseudo-randomized y-coordinates of the scatterplots. The lines shown in the validation task had random deviations within $\pm [0.0,1.0]$ from  the true mean of the data points.

\textbf{Results}: The errors in the estimation tasks were roughly normally distributed with $\mu=0.02$ and a maximum deviation of $0.7$. Lines shown in the validation tasks with deviations within the 95\% CI ($dev<0.198$) had acceptance rates of at least 80\%, while every line with $dev>0.198$ had an acceptance rate of at most 30\%. Overall, participants were more accurate in visual estimation than visual validation, and they exhibited a slight difference in judgment between positive and negative deviations in the validation tasks. Most participants reported using a ``counting'' strategy to solve the tasks, approximating the number of points above and below a given line.

\begin{figure}
 \centering
 \includegraphics[width=0.7\columnwidth]{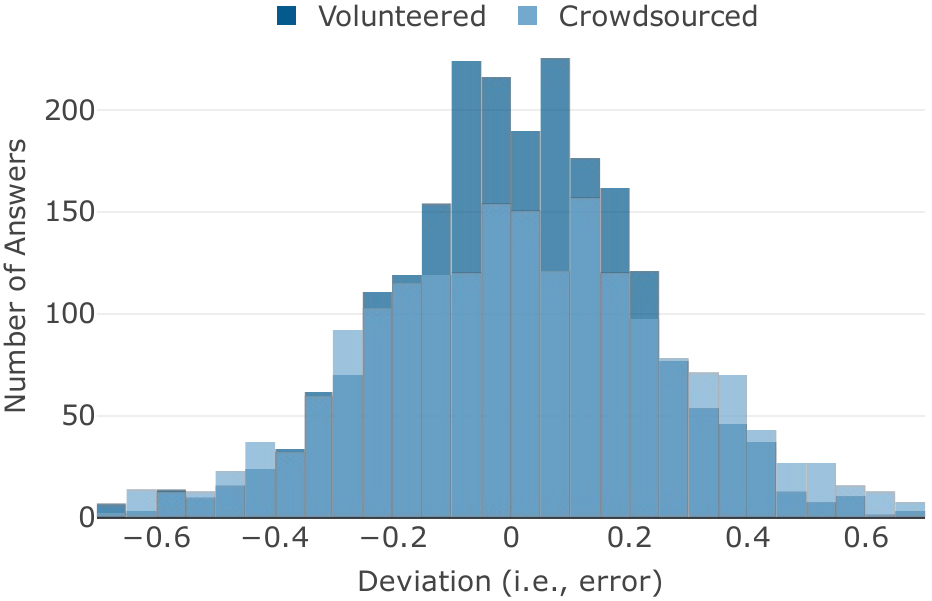}
 \caption{Histogram of the deviations of the drawn lines in the estimation tasks for both populations.}
 \label{fig:est_error}
\end{figure}

\subsection{Main Study}
\label{ssc:mainstudy}

Analysis of the pilot study's results led to the hypotheses:
\begin{itemize}[noitemsep, nolistsep]
    \item \textbf{H1}: The accuracy of visual validation is lower than the accuracy of the visual estimation when perceiving the mean value of points in a scatterplot.
    \item \textbf{H2}: People's critical point between accepting and rejecting a given mean value is close to the boundary of the 95\% CI.
    \item \textbf{H3}: For visual validation, the results differ between positive and negative deviation from the true mean.
\end{itemize}

\subsubsection{Stimuli Design}
\label{sssc:stimulidesign}

Based on the findings from the pilot study and for matching the distribution assumption of linear regression residuals, the y-coordinates of the points were generated from a normal distribution with random mean between 30 and 70 and standard deviation between 15 and 25. 
Following the literature\cite{Gleicher.2013, Hong.2022}, we focused on the perception of only one dimension (i.e., the y-axis). The adoption of a normal distribution is prevalent across numerous applications (e.g., as a pre-requisite of least square regression) and provides consistent conditions throughout all trials, given that the pilot study results were partially influenced by the stimulus. Thus, the level of trial difficulty was determined by the deviation of the displayed line in the validation tasks.
Since most participants approximated the median instead of the mean in the pilot study, we made sure that the number of points above and below the true mean differed by at least 10\% to discourage the use of ``bounding boxes''\cite{Yang.2019} as a perceptual proxy. 

The pilot study showed that lines with deviations greater than $0.7$ were consistently rejected. Thus, we used lines with evenly distributed deviations in the range of $\pm [0.0,0.7]$ to determine participants' acceptance threshold. Consistent with previous work\cite{Yang.2019}, logistic regression was chosen to analyze the validation task (see \autoref{sssc:analysis}). A power analysis of the logistic regression of the pilot study indicated that a sample size of at least 50 was necessary to obtain a meaningful model\cite{Motrenko.2014}. Therefore, the study included 50 trials with 25 lines with a positive and 25 lines with a negative deviation in the validation task.

\begin{figure}
    \centering
    \begin{subfigure}[t]{0.65\columnwidth}
         \centering
         \includegraphics[width=\columnwidth]{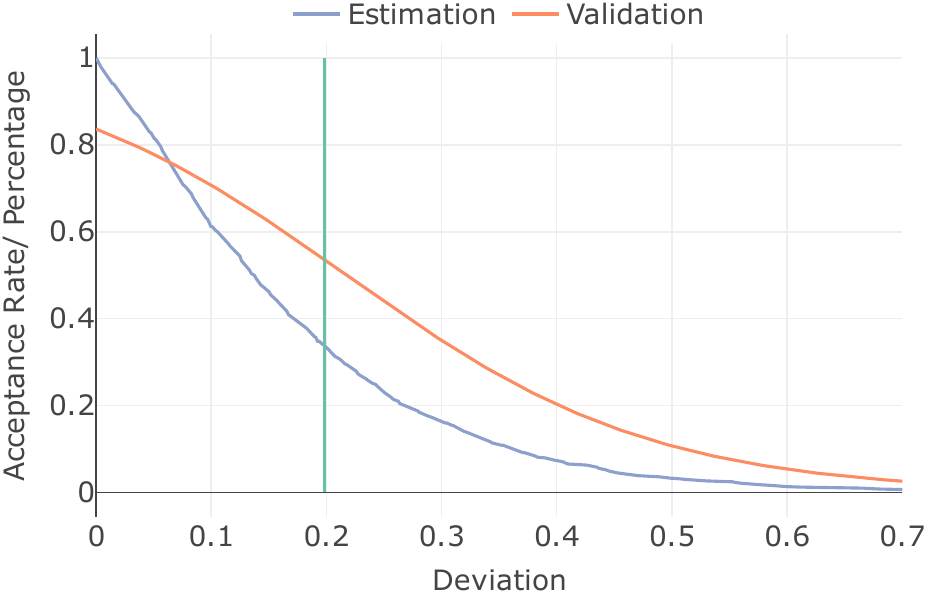}
         \caption{Volunteered population.}
         \label{fig:comp_normal}
     \end{subfigure}
    
     \begin{subfigure}[t]{0.65\columnwidth}
         \centering
         \includegraphics[width=\columnwidth]{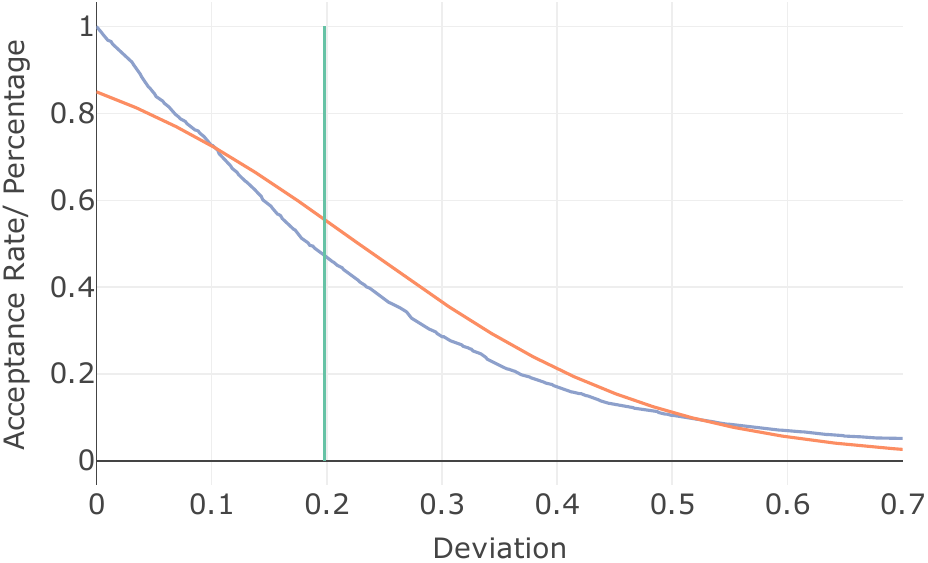}
         \caption{Crowdsourced population.}
         \label{fig:comp_prolific}
     \end{subfigure}
        \caption{Validation and estimation accuracy for both populations (absolute deviation). Blue line: Cumulative distribution for the estimation errors. Orange line: Logistic regression for the validation acceptance. Green line: Statistical 95\% CI.}
        \label{fig:comp}
\end{figure}

\subsubsection{Experimental Setting \& Participants}
\label{sssc:expsetting}

To increase the generalizability of our findings, we conducted an online study with two different populations. The first population consisted of 100 individuals who volunteered to participate after seeing advertisements in university lectures and mailing lists. After filtering out 14 participants who failed attention checks, 42 individuals completed the validation and 44 the estimation task.
For the second population, we recruited 90 participants via crowdsourcing platform Prolific\cite{Prolific}. After filtering for attention checks, 42 individuals completed the validation and 40 the estimation task.

In both populations, most participants were between 20 and 30 years old (43\% for volunteered, 72\% for crowdsourced) and nearly evenly split between women and men (volunteered: 46\%~F, 50\%~M, 4\%~other; crowdsourced: 47\%~F, 52\%~M, 1\%~other). The level of education and experience with statistical model estimation was slightly higher among the volunteers.

\subsubsection{Analysis}
\label{sssc:analysis}

In the validation task we measured whether participants accepted the displayed line as the true means. 
To ensure comparability with the estimation task results, we transformed the responses to binary results.
Logistic regression was then applied to the acceptance rates of the shown lines, which is a technique that has been used in previous work\cite{Yang.2019}.
In the estimation task, we measured whether participants were able to draw (estimate) the true means. 
The estimation errors were measured as the deviation of the lines drawn by the participants.

For statistical testing, we first ran a Shapiro-Wilk test on the given responses and response times to see if they were normally distributed. Although none of the tests were positive, a visual inspection of the acceptance rates and estimation errors suggested that this was due to the large sample size (see \autoref{sssc:results}). Therefore, we used t-tests to compare these results. We used a Kruskal-Wallis test for the response times and a chi-squared test for comparing Likert responses. 

\subsubsection{Results}
\label{sssc:results}

\paragraph{Distribution and Bias}

\begin{figure}
 \centering
 \includegraphics[width=0.7\columnwidth]{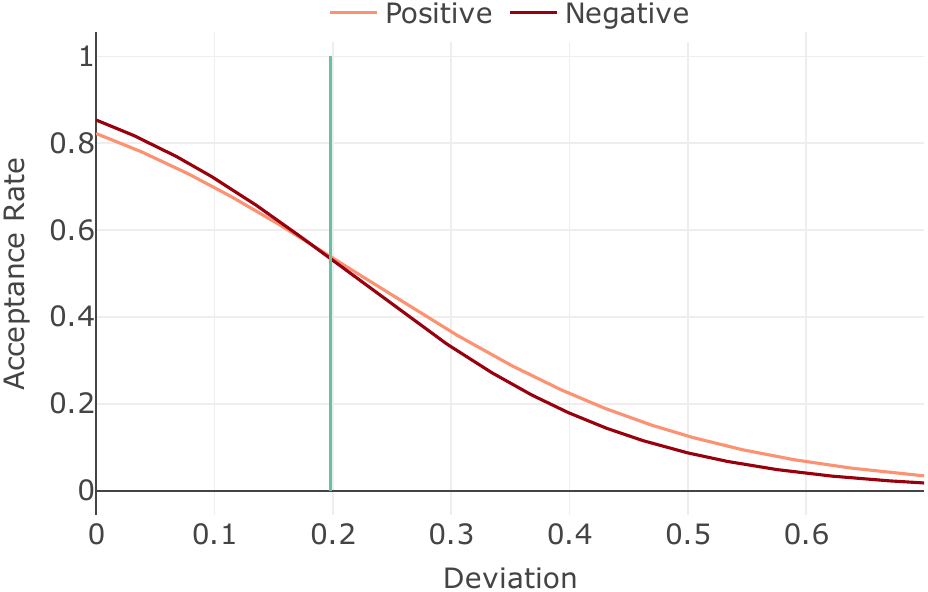}
 \caption{Comparison of the logistic regressions for the acceptance rates of positive and negative deviations of the volunteered population. Green line: statistical 95\% CI.}
 \label{fig:val_error_comp}
\end{figure}

For both populations, the estimation errors resembled a normal distribution (\autoref{fig:est_error}; $\mu_{volun}<0.01$, $\mu_{crowd}=0.06$), consistent with the regression assumptions. The same is true for the acceptance rates of the lines displayed for validation ($\mu_{volun}=0.01$, $\mu_{crowd}<0.01$). The means of the errors/ acceptance rates being zero indicate that neither of the processes is biased. This finding is supported by a comparison of the acceptance rates for positive and negative deviations from the mean (\autoref{fig:val_error_comp}), which showed no significant difference ($\mbox{p}_{volun}=0.29$, $\mbox{p}_{crowd}=0.72$).
This rejects our hypothesis \textbf{H3}.

\paragraph{Accuracy of Visual Validation vs. Visual Estimation}

Since both estimation errors and acceptance rates were approximately normally distributed, we considered positive and negative deviations cumulatively as absolute deviations. We compare the logistic regression for the acceptance rates of the validation task with the cumulative distribution of the estimation errors in \autoref{fig:comp}. 
For the volunteered population, the acceptance threshold for validation was less accurate than the visually estimated ones ($p<0.01$, $\mbox{cohensD}=0.35$).
For the crowdsourced population, the logistic regression for validation was almost identical ($p=0.51$), but the estimation errors were higher.
Although the accuracy of the visually accepted lines was lower than the accuracy of the visually estimated lines, the difference was not significant ($p>0.29$).
In summary, \textbf{H1} is partially accepted.
  
\paragraph{Critical Point of Validation}

\begin{figure}
 \centering
 \includegraphics[width=0.7\columnwidth]{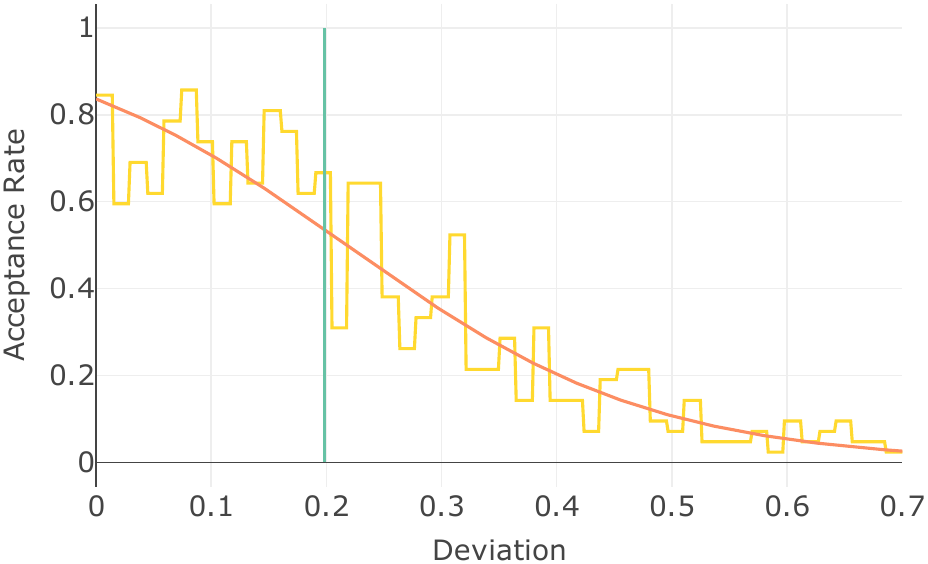}
 \caption{Acceptance rate of the shown lines in the validation tasks (absolute deviation) of the volunteered population. Yellow line: The raw percentages per deviation. Orange line: The logistic regression of the yellow line's data. Green line: Statistical 95\% CI.}
 \label{fig:val_error}
\end{figure}

\autoref{fig:val_error} shows the raw acceptance rates per deviation and the corresponding logistic regression exemplary for the volunteered population. The inflection points of the logistic regressions correspond by construction to the 50\% acceptance rate. Their values $dev_{volun}=0.217$ and $dev_{crowd}=0.228$ were very close to the boundary of the statistical 95\% CI ($dev=0.198$). Moreover, the critical points of the individuals' logistic regressions resembled a normal distribution near the 95\% CI ($\mu_{volun}=0.223, \mu_{crowd}=0.241$). This indicates that people's perceived visual confidence interval matches the statistical 95\% CI and accepts hypothesis \textbf{H2}.

\paragraph{Difficulty and Response Time}

For both populations, no significant difference in response time (\textit{Wilcoxon-test:} volunteered: $\mbox{p-value}=0.94$, $\mu_{val}=11.3\mbox{sec}$, $\mu_{est}=11.8\mbox{sec}$; crowdsourced: $\mbox{p-value}=0.50$, $\mu_{val}=10.2\mbox{sec}$, $\mu_{est}=10.5\mbox{sec}$) and reported difficulty (\textit{chi-squared test:} volunteered: $\mbox{p-value}=0.26$, $\mu_{val}=2.76$, $\mu_{est}=3.16$; crowdsourced: $\mbox{p-value}=0.43$, $\mu_{val}=3.14$, $\mu_{est}=3.22$) was found between the two tasks.

\paragraph{Self-Reported Strategies}

For visual estimation, most participants derived the mean using a perceptual proxy and ``adjusted it for outliers" (without having true statistical outliers, the participants probably meant points that were a bit off). These perceptual proxies were based on the density of the scatterplot, the median, or the distance between the highest and lowest points.

In the visual validation tasks, the majority of the participants validated the line based only on the number of points above and below the line. However, some persons judged based on the perception of the density and structure of the points. Six participants ``estimated [their own] mean and compared it to the shown line.''

\section{Limitations and Future Work}
\label{sc:fw}

The participants in this study were non-experts in data visualization or statistics, making them not representative of domain experts.
We chose to focus on a specific type of data(-distribution), visualization, and model as a starting point to understand visual model validation. Although our findings are statistically sound, the generalizability of our results regarding these aspects remains to be established.

Furthermore, we need to investigate the perceptual mechanisms involved in performing visual model validation, and how they compare to the mechanisms of visual model estimation. 
By imposing a time limit during the trials, participants would likely stop relying on ``counting strategies'', but instead adopt other perceptual proxies for solving the tasks. It would also be interesting to explore the influence of visual encoding and data patterns (e.g., outliers, shapes), as well as dimension size and number of points. By doing so, we could derive design guidelines for model visualizations that mitigate perceptual biases in visual validation (e.g., the work by Hong et al.\cite{Hong.2022}) and provide prescriptive instructions on when and how visualizations should include pre-drawn models fitted to data.

\section{Conclusion}
\label{sc:conclustion}

Our empirical user study with two different populations investigated the difference between visual model estimation and visual model validation by using the average value in scatterplots as a baseline. Our findings suggest that visual model validation accepts models that are less accurate than those that are estimated visually. The acceptance level is similar to the common statistical standard 95\% confidence interval. 
Our study provides valuable insights into how humans process statistical information and identifies limitations and potential aspects for future research.

\acknowledgments{
The authors would like to thank all study participants and the reviewers, whose suggestions helped to improve this paper. This paper is a result of Dagstuhl Seminar 22331 "Visualization and Decision Making Design Under Uncertainty". This work has been partially supported by BMBF WarmWorld Project and Risk-Principe Project. This work has been funded in part by NSF Awards 2007436, 1452977, 1940175, 1939945, 2118201.}

\bibliographystyle{abbrv-doi-hyperref}

\bibliography{literature}

\end{document}